\documentclass[conference]{IEEEtran}
\usepackage{graphicx}
\usepackage{color,soul}
\usepackage{algorithm}
\usepackage{algorithmic}
\usepackage{amsmath,amssymb}
\usepackage{subcaption}
\usepackage{enumitem}
\ifCLASSINFOpdf
\else
\fi
\begin{document}
%
\title{Density-Based Pruning of Drone Swarm Services}


\author{\IEEEauthorblockN{Balsam Alkouz}
\IEEEauthorblockA{\textit{School of Computer Science} \\
\textit{The University of Sydney}\\
Sydney, Australia \\
balsam.alkouz@sydney.edu.au}
\and
\IEEEauthorblockN{Athman Bouguettaya}
\IEEEauthorblockA{\textit{School of Computer Science} \\
\textit{The University of Sydney}\\
Sydney, Australia \\
athman.bouguettaya@sydney.edu.au}
\and
\IEEEauthorblockN{Abdallah Lakhdari}
\IEEEauthorblockA{\textit{School of Computer Science} \\
\textit{The University of Sydney}\\
Sydney, Australia \\
abdallah.lakhdari@sydney.edu.au}
}


%


\maketitle

\begin{abstract}
We propose a novel framework for the recommendation of swarm-based drone delivery services based on the consumers preferences. We propose a density-based pruning approach that uses the concept of partnerships with charging station providers to reduce the search space of swarm-based drone service delivery providers.
A weighted service composition algorithm is proposed that considers the providers capabilities and consumers' preferences in selecting the best next service. We propose a voting-based recommendation algorithm to select the best providers. We conduct a set of experiments to evaluate the efficiency of the framework in terms of consumer satisfaction, run-time, and search space reduction cost.
\end{abstract}


%
\IEEEpeerreviewmaketitle

\section{Introduction}
Drone \textit{delivery services} are the new technology that promises impactful and innovative solutions to the ever expanding online shopping \cite{khan2018drones}. COVID-19 pandemic has provided a unique opportunity for the deployment of drones as an emerging and most preferable alternative for service delivery \cite{Golden2020Age}.
Key benefits include the ability to deliver goods in a socially distanced and contactless manner and ensuring a more resilient supply chain.  The pandemic has also highlighted that the benefits of drone-based delivery far outweighs any potential risks, especially  in last-mile deliveries \cite{Golden2020Age}.
Big players have expanded their drone delivery services recently with Google's Wing doubling their deployment rates. Their partners have claimed a 50\% increase in sales during the crisis\footnote{Drone delivery golden age: https://www.weforum.org/agenda/2020/07/golden-age-drone-delivery-covid-19-coronavirus-pandemic-technology/}. \looseness=-1

Drone swarm services are a means of increasing the delivery capabilities of a single drone \cite{alkouz2020swarm}. A consumer ordering multiple packages that exceed the capability of a single drone would necessitate the use of a drone swarm. Large drones, on the other hand, are prohibited in cities\footnote{https://www.faa.gov/uas/advanced\_operations/package\_delivery\_drone}, mandating the adoption of multiple smaller drones cooperating together to deliver packages \cite{alkouz2021provider}. Therefore, a swarm of drones is needed to deliver packages to the \emph{same destination at the same time}. Furthermore, a swarm is required when the energy consumption burden, due to the payload, needs to be shared to allow the swarm travel further distances \cite{alkouz2020formation}.

Drone swarm delivery services play a big role in the future of smart cities \cite{abdelkader2021aerial}. 
Drones in a swarm may need to stop at recharging or delivery stations located on building rooftops \cite{alkouz2020swarm}. These building rooftops represent the nodes in a skyway network \cite{shahzaad2020game}. The Line of Sight (LoS) paths between the nodes are the skyway segments. \emph{We abstract a swarm carrying packages and travelling in a skyway segment as the swarm-based drone delivery service,} known as Swarm-based Drone-as-a-Service (SDaaS). In this respect, the functional part of the service represents the delivery of packages from a node to the next LoS node.
The non-functional (QoS) aspects of the service include delivery time, cost, energy consumed, etc. 

Integrating swarm-based delivery services into supply chains requires coordination between the service providers and consumers. Majority of work on swarm-based delivery services focus on the provider perspective, i.e increasing the profit or reducing the delivery time to reallocate the drones to other requests \cite{alkouz2021reinforcement}. However, there is a gap in work done on drone swarm services composition from a consumer perspective. A consumer would typically have preferred quality of services, i.e. cost, delivery time, etc. The services composition would need to be customised to satisfy the consumer's needs. In addition, previous works on composition focus on a single objective optimization, eg: reducing delivery time \cite{alkouz2020swarm}. In this work, we propose to compose the path with a multi-objective optimization of the consumers QoSs. Therefore, In this work, we propose the concept of \textit{swarm-based services broker} that works on behalf of the consumers. A broker is a 3rd party that acts as an intermediate between consumers and providers. Providers would typically want to sign up with the broker for more consumer exposure.

There are several delivery service providers that deliver packages from  warehouses, shops, and restaurants. An example of conventional food delivery services include Uber eats, Menulog, Didi, Zomato, etc. In the same manner, we envision the presence of multiple swarm-based delivery service providers that compete to maximize their profit margins. Hence, a \emph{recommendation mechanism} is required from the broker to recommend the best provider for a request. On the other hand, a consumer has certain weighted preferences for the different QoSs  \cite{wang2009web}. The recommendation of the provider is dependant on the swarm-based delivery \emph{services composition} that composes the best services from a source to a destination. Note that a service is defined as a swarm travelling on a skyway segment. Hence, the broker needs to compose the best skyway segments to the destination with the optimal \textit{preferred} QoSs. Composing recommending services for all available providers is a computationally expensive task. Therefore, the broker's algorithm needs to reduce the search space to providers that are most likely going to satisfy the consumer. The goal of this paper is to \textit{reduce the search space}, \textit{compose the best services}, and \textit{recommend the best service providers} for a consumer's request. The composition and recommendation consider the consumers preference in terms of cost, delivery time, and the rest of QoS attributes.

We define our main contributions as follow:
\begin{itemize}[leftmargin=*]
    \item A novel preference-based Swarm-based Drone-as-a-Service (SDaaS) framework.
    \item A density-based pruning approach to reduce the search space of SDaaS providers.
    \item A weighted SDaaS services composition algorithm to compose the best services given the consumer's preferences, the providers capabilities and techniques, and the environment constraints and partnerships.
    \item A voting-based recommendation of SDaaS providers.
    
\end{itemize}

\section{Motivating Scenario}
\begin{figure}[htbp!]
\centering
\includegraphics[width=\linewidth]{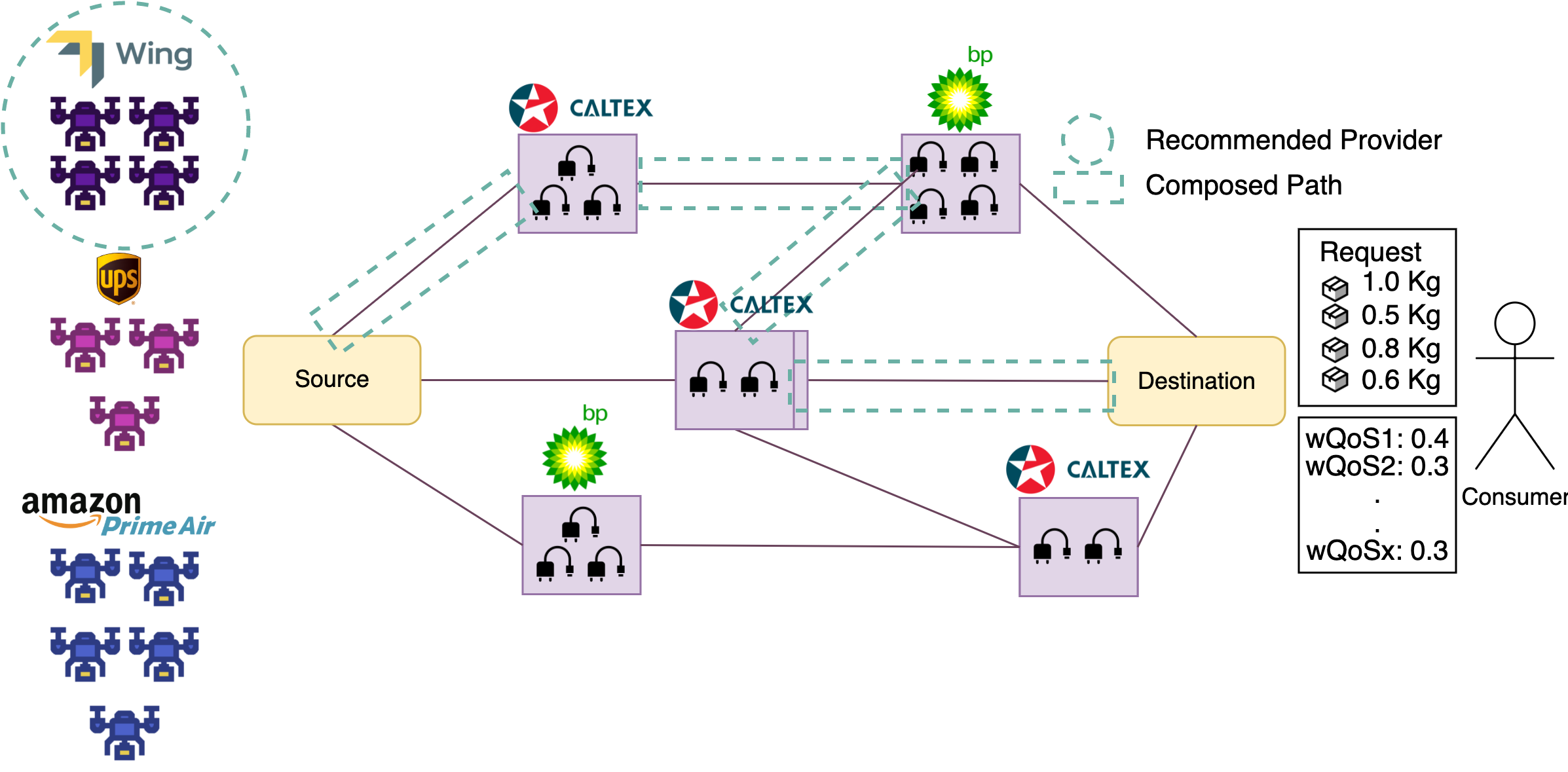}
\caption{Motivating Scenario}
\label{fig:motivating}
\end{figure}

Let us assume a hospital requests multiple pieces of medical equipment to be delivered from a medical supplier warehouse. The equipment pieces need to be delivered together as fast a possible. A swarm of drones is therefore required to deliver these equipment pieces as they exceed a single drones payload capacity. We assume that there are multiple delivery service providers available to deliver the equipment (see Fig. \ref{fig:motivating}). Each swarm is assumed to have different capabilities which includes different payload capacities, battery capacities, speeds, and number of drones. Furthermore, we assume that each swarm consists of homogeneous drones, i.e. drones of the same type and capabilities. Moreover, we assume no package handover occurs, i.e. the packages are carried by the same set of drones from the source to the destination.

A sky network is provided to facilitate the smooth and efficient delivery of packages from source to destination. The skyway network consists of rooftops (i.e. network nodes) equipped with recharging stations. We assume that these recharging stations belong to different recharging service providers. We assume that all charging stations have pads with similar charging speeds. The skyway segments are the Line of Sight (LoS) paths between nodes. Some providers have partnerships with recharging station providers. An example in the transport service sector is the partnership between Uber and Caltex petrol service stations that gives discounts to Uber drivers to refuel their cars\footnote{Caltex fuel discounts: https://www.uber.com/en-AU/blog/fuel-discount-terms-and-conditions/}. Partnerships in our scenario are means to prioritize the use of recharging stations which in return reduces the waiting times due to occupied stations. In addition, partnerships allow for reduced costs on swarm service providers when recharging. We assume that each SDaaS provider has partnership with one charging provider.

The delivery environment is constrained by intrinsic and extrinsic constraints \cite{jermaine2021demo}. The intrinsic constraints include the different payloads carried between the drones which causes difference in power consumption. They also include the different power consumption rates due to the drones positions in a swarm formation affected by drag and upwash/downwash forces \cite{alkouz2020formation}. Extrinsic constraints include the availability and number of charging stations at different nodes. They also include the changing wind conditions. These constraints need to be taken into consideration when designing a swarm-based delivery system. The delivery time is typically affected by the drones speed, the charging time at intermediate nodes, and the waiting time due to sequential charging if the size of the swarm is larger than the available recharging stations.

The goal is to \emph{compose the optimal path from the source to the destination for providers, given the various constraints, and recommend the best provider for a delivery request}. Each provider has their own techniques that may be used to optimize their deliveries. These include the use of flight formations that adapt to various wind conditions \cite{alkouz2020formation}. They also include a swarm's ability to split and merge, reducing congestion at intermediate nodes \cite{alkouz2020swarm}.
As a result of the partnerships, swarm capabilities, and techniques employed, \textit{the service composition for the various providers would differ}.

On the other hand, a consumer (the hospital) may weight the QoS attributes in order of preference. For example, in the case of an emergency, the delivery time outweighs the cost and energy consumption QoSs'. The service composition should be \textit{customized to serve the consumer preferences} and weighted attributes \cite{yu2012multi}. Therefore, we propose a \textit{swarm-based services broker} that acts on behalf of the consumers to compose the optimal services for every provider. The broker \emph{composes the optimal path for service providers} based on the consumers preferences and the providers capabilities under the different constraints. Composing the path for all $n$ providers would be computationally costly. Therefore, a \textit{search space reduction} is required to only compose the path for the most relevant providers for the consumers request. The broker \emph{recommends the best matching provider for a given request} taking the consumer's preferences and expectations into consideration.

\section{Related Work}

The study of swarm-based applications has been widely explored in the literature. Drone swarms are being used in target search, airborne communication, skyshows, etc \cite{alkouz2021provider}. 
Majority of the work that discuss drone swarms in delivery refer to them as multiple independent coordinated single drones used to deliver packages to multiple destinations \cite{kuru2019analysis}. In contrast, we define a swarm as a set of drones that act as a single entity and move together to deliver multiple packages to the same destination. Swarm-based Drone-as-a-Service (SDaaS) is defined as the abstraction of using a drones swarm to deliver packages between two nodes in a skyway network \cite{alkouz2020swarm}. In a real-life scenario, multiple SDaaS service providers compete to serve a consumer's request to boost their revenues. Majority of work on SDaaS allocate and compose services from a provider perspective with a single objective optimization criteria, e.g. delivery time or profit \cite{alkouz2020swarm}\cite{alkouz2021provider}. To the best of our knowledge, this work is the first to discuss SDaaS services from a \textit{multi-objective consumer perspective}. This work designs \textit{tailored} compositions based on every consumer preferences.

Service provider recommendation criteria are of major importance to the service industry. Services, unlike goods, cannot be evaluated prior to purchase \cite{kugyte2005standardized}. The factors that affect the recommendation of a provider can be classified to two main groups. The first is the delivery of service. This includes pricing, inconvenience, service failures, and competition \cite{kugyte2005standardized}. The second is the representatives personal characteristics. This includes service encounter, ethical problems, and empathy \cite{kugyte2005standardized}. Since swarm-based drone delivery services are considered a mass service where there is no formal relationship with the consumer \cite{kugyte2005standardized}, \emph{we focus on the first group of factors that deal with the delivery of service}. The consumers may provide their QoS expectations and service providers may also express their offers. Hence, a dynamic framework is required to take the consumer requirement as an input and provide the best provider as an output. Due to the possibility of a high number of providers, a \textit{search space reduction} strategy is required to compose services for just potentially competitive providers.

The research of search space reduction in drone services has primarily relied on spatial data from competing providers \cite{shahzaad2020game}. In our scenario, this strategy is not viable because we presume that all providers considered are inside a spatially reachable area. Other research on search space reduction has only focused on reducing the number of nodes examined in routing within a network \cite{guo2003search}. Other recommendation systems rely solely on screening out ineligible providers who are unable to meet the consumer's request. One example is hotel booking systems, which provide all providers who can meet the consumer's needs \cite{malviya2019web}. Because our system would need to compose the path for each provider before recommending, our search space reduction strategy should concentrate on lowering the number of providers considered. Following the reduction and composition, the ideal provider would be recommended. One method to recommend the best provider to a consumer requirement is using a voting-based system \cite{baranwal2014framework}.

Voting systems are generally associated with political elections. A voting system is a set of rules that regulate how results of elections are determined \cite{gallagher2005politics}. Of particular interest, is how ballots are counted. Ranked voting is a system where voters rank their choices in order of preference \cite{obata2003method}. Ranked voting has been used to select the best cloud service provider \cite{baranwal2014framework}. 
However, different vot count systems have not been used in provider recommendation scenarios. These systems include Plurality, Borda Count, Condorcet, etc \cite{poundstone2008gaming}. Hence, we propose to implement a voting-based service provider recommendation system that utilizes multiple voting systems. 
In addition, to the best of our knowledge, there is no work done on swarm-based services providers recommendation to recommend the best provider for a consumer's request. Therefore, this work is the first to propose a \textit{swarm-based drone delivery recommendation system based on consumers needs}. 

\section{Multi-provider Swarm-based Drone-as-a-Service Model}
\begin{figure}[!t]
\centering
\includegraphics[width=0.8\linewidth]{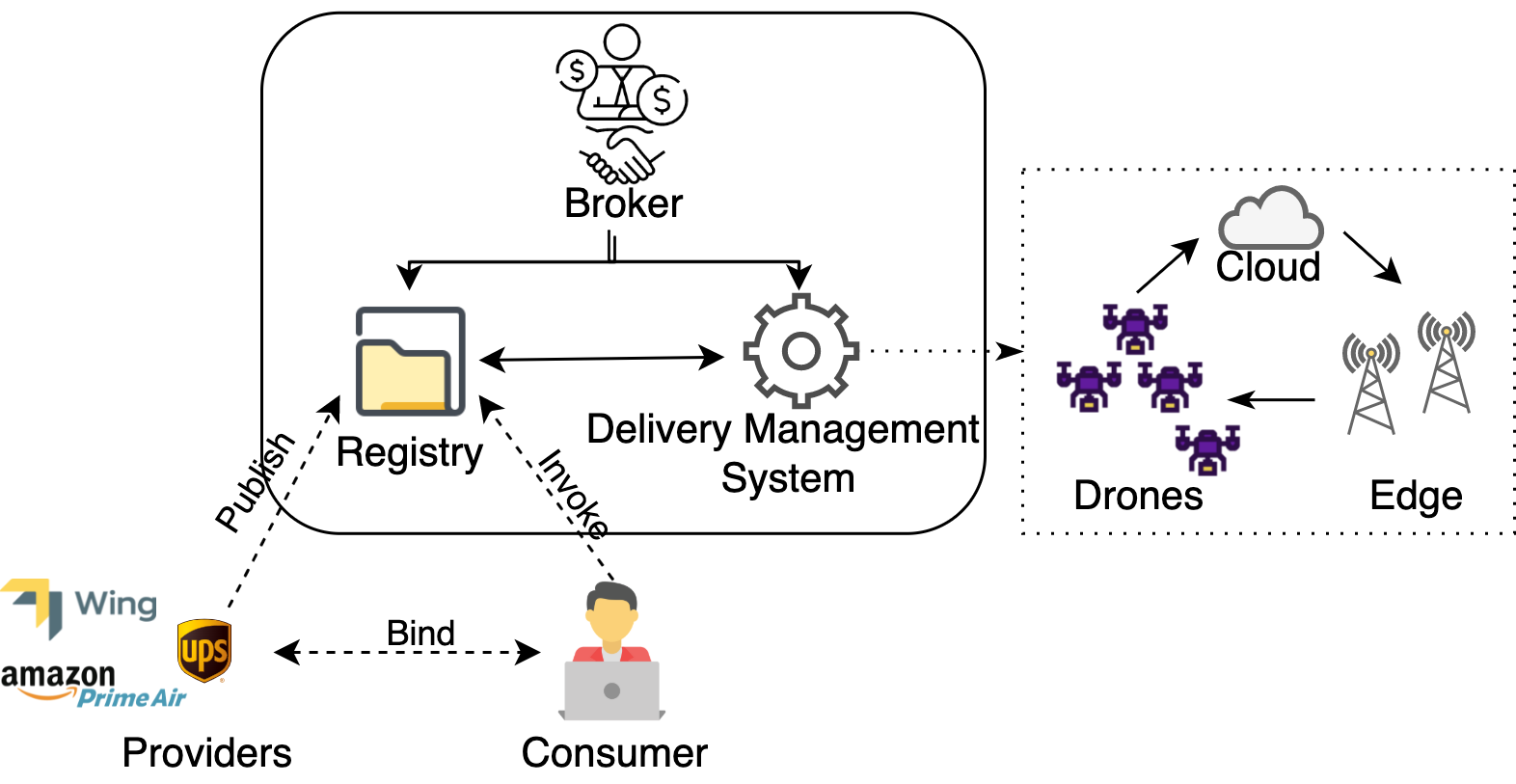}
\caption{Multi-provider SDaaS System Architecture}
\label{fig:architecture}
\end{figure}

The architecture of the Multi-provider SDaaS is premised on having multiple providers competing to serve a consumer's request. The consumer and providers' trusted \emph{broker} eliminates noncompetitive providers, composes the paths, and recommends the best provider for the consumer's request. The architecture is an adapted SOA architecture \cite{de2011building}. The providers register their services in the registry stating all the capabilities of their swarm. We assume that the integrity of the provided information is ensured. The consumers on the other hand provide their preferred QoS, delegates the broker to recommend the best providers, and invokes the services. The delivery management system, managed by the broker, \emph{composes the paths and recommends the best matched service provider} for the consumer. The delivery management system, in addition, coordinates between the selected swarm, cloud, and edge nodes. The \emph{swarm} as it traverses the network publishes its location and battery states to the cloud. The \emph{cloud} updates the information on the edge nodes which are placed in strategic locations in the city. The \emph{edge} nodes send instructions to the swarm on where to go next (see Fig. \ref{fig:architecture}). 

In this section, we formally define SDaaS providers, SDaaS, and consumers' requests.

\textbf{Definition 1: SDaaS Service Provider.} An SDaaS provider is defined as a tuple of $<Provider\_id,S>$
\begin{itemize}[leftmargin=*]
    \item $Provider\_id$ is a unique provider identifier. There could be $1..n$ providers, where $n$ is the maximum number of providers able to serve a request.
    \item $S$ is the swarm of a provider that is a core component of an SDaaS. Providers may own a different number of delivery drones $D_d$, have different payload capabilities $p$, different battery capacities $b$, and exhibit different techniques $t$.  
\end{itemize}

\textbf{Definition 2: Swarm-based Drone-as-a-a-Service (SDaaS).} An SDaaS is defined as a tuple of $<SDaaS\_id, S, F, QoS>$, where
\begin{itemize}[leftmargin=*]
    \item $SDaaS\_id$ is a unique service identifier
    \item $S$ is the swarm travelling in SDaaS. $S$ consists of $D$ which is the set of delivery drones forming $S$, a tuple of $D$ is presented as $<D_{d1},..,D_{dz}>$. $S$ also contains the properties including battery levels of every $d$ in $D$ $<b_1,..,b_z>$, the payloads that every $d$ in $D$ is carrying $<p_1,..,p_z>$, and the techniques that $S$ exhibit $<t_1,..,t_y>$. 
    \item $F$ describes the delivery function of a swarm in a skyway segment between two nodes, A and B. $F$ includes the travel, charging, and waiting times of the segment.
    \item $pQoS$ are the quality of services perceived for every service served by $S$. This includes the delivery time $dt$, energy consumed $ec$, execution time $et$, travel cost $c$, etc. 
\end{itemize}

\textbf{Definition 3: SDaaS Request.} A consumer request is a tuple of $<\alpha, \beta, P>$. $\alpha$ is the source node, $\beta$ is the destination node, $P$ are the weights of the packages requested where $P$ is $<p_1,p_2,..,p_l>$, and $wQoS$ are the weighted consumer's preferred QoSs where $wQoS$ is $<wQoS_1,wQoS_2,..,wQoS_x>$.\\

\textbf{Definition 4: SDaaS Provider Charging Partnership.} A partnership between an SDaaS provider and a charging station provider would entitle benefits like reduced costs and priority charging. A partnership is defined as a tuple of $<Provider\_id, Station\_id, Partnership\_tier>$, where
\begin{itemize}[leftmargin=*]
    \item $Provider\_id$ is the ID of the SDaaS service provider, i.e the first party in the partnership.
    \item $Station\_id$ is the ID of the charging station provider, i.e. the second party in the partnership.
    \item $Partnership\_tier$ is the level of partnership that would define the cost and priority benefits, e.g. \textit{platinum, gold, and silver}. Each tier would have a fixed cost for charging per kWh\footnote{https://www.carsales.com.au/editorial/details/what-does-it-cost-to-recharge-an-ev-119255/}. The priority is defined by how much the swarm would wait before accessing a charging pad in case of congestion. For example, a platinum partnership may entitle the swarm to skip the whole queue waiting to recharge while the gold membership would allow 2 drones in the queue to be served before. Skipping queues for reward programs members is common. For example, first and business class passengers skip the queue boarding an airplane\footnote{https://stampme.com/the-power-of-vip-perks-and-exclusive-benefits-in-rewards-programs/}.
\end{itemize}

\subsection{Service Providers Techniques}
\label{techniques}
There are multiple \emph{techniques} a service provider may utilize to improve its QoS. Such techniques include the behaviour of the swarm, formation of the swarm, and cooperation. These techniques are inspired by other works proposed on SDaaS services composition \cite{alkouz2020swarm}\cite{alkouz2020formation}. The broker, based on the providers chosen techniques, will compose the path accordingly. Below we describe each technique in details.
\begin{itemize}[leftmargin=*]
    \item \emph{Swarm Behaviour}: A swarm may inhibit different behaviours including static and dynamic organization. A \emph{static} behaviour is when the swarm members are formed at the source and stick together throughout the delivery journey \cite{akram2017security}. The benefit of this method is ensuring that all the packages would arrive to the destination at the same time. In contrast, a \emph{dynamic} swarm is a swarm whose members are decided at the source but whose structure may change throughout the journey \cite{akram2017security}. The swarm in this behaviour may band and disband in the network \cite{alkouz2020swarm}. The benefit of this method is that it reduces the time spent at recharging stations due to sequential charging since the swarm disbands and disperses in the network avoiding congestion.
    \item \emph{Swarm Formation}: A swarm may be shaped into different formations. This includes Vee, Diamond, Front, Echelon, and Column formations. Different formations are optimal in terms of energy consumption under different wind conditions \cite{alkouz2020formation}. A \emph{fixed} formation is when the shape of the swarm is decided at the source and stays the same throughout the journey. A \emph{flexible} formation, is when the swarm shape changes whenever the wind changes. This adaptivity would allow the swarm to consume less energy but may be computationally expensive \cite{alkouz2020formation}.
    \item \emph{Cooperation}: Drones in a swarm may cooperate on their own expense and recharge less at intermediate nodes to reduce the charging times \cite{alkouz2020swarm}. Drones in a cooperative swarm charges up to what takes it to the next neighbouring nodes instead of recharging fully.
\end{itemize}

\section{SDaaS Composition and Provider Recommendation Framework}
\begin{figure*}[!t]
\centering
\includegraphics[width=0.95\linewidth]{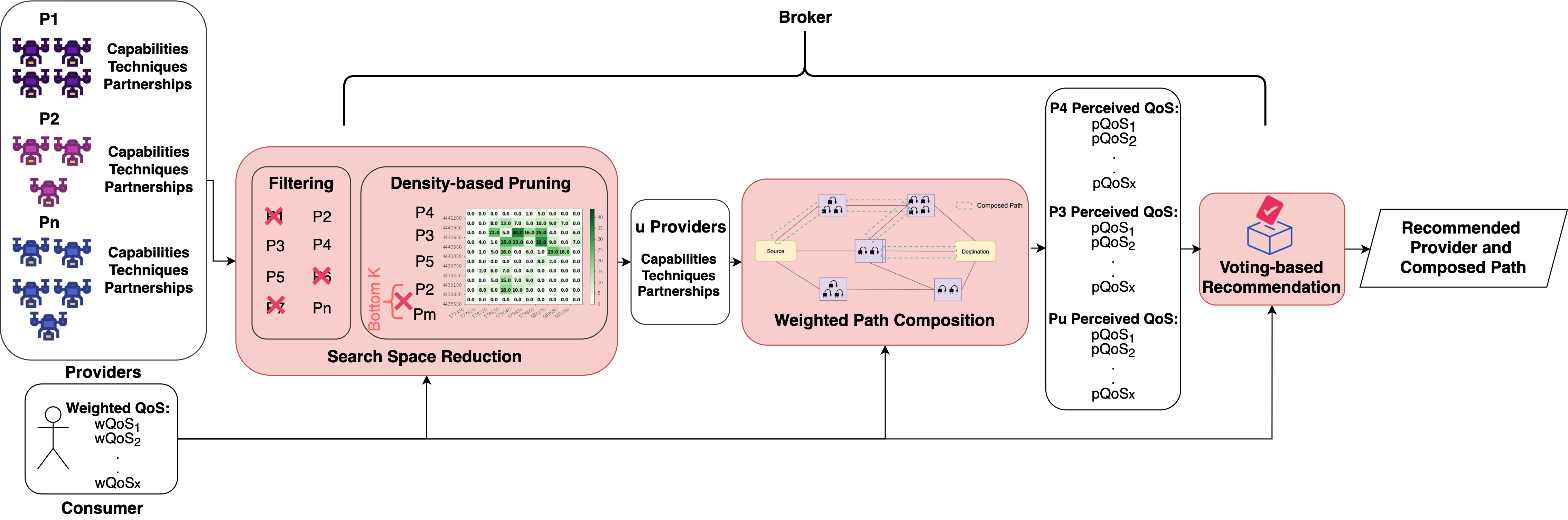}
\caption{SDaaS Recommendation Framework}
\label{fig:framework}
\end{figure*}

In this section, we first describe the QoS metrics that characterize consumers preferences. Then, we discuss the three main steps involved in the proposed SDaaS recommendation framework (Fig. \ref{fig:framework}). First, is the search space reduction to reduce the number of service providers considered in the composition and recommendation. Second, is the optimal composition of the path for every considered provider. The third is the recommendation of the best service providers. In this section we describe these processes taking the consumer preferences in consideration.

\subsection{SDaaS QoS Metrics}
\label{qos}
We define the different QoS metrics that characterize consumers preferences and delivery service qualities by providers. These metrics are key to achieving an optimal preference-based service recommendation of the providers. Unlike tangible goods, the measurement of a service quality is non-direct. Hence, scientists introduced several research instruments to help measure the quality of a service. Of those instruments, SERVQUAL is designed to capture consumers expectations and perceptions along five dimensions \cite{parasuraman1988servqual}. SERVQUAL is also known as RATER, which stands for the five dimensions or service factors it measures. This includes reliability, assurance, tangibles, empathy and responsiveness. 
In this paper we focus on the tangibles, reliability, and responsiveness as they are directly measured by the composition of services in the path composition step in subsection \ref{composition}. Assurance and empathy do not apply to the mass SDaaS services where there is no formal relationship with the consumer. These factors will be used to recommend the best SDaaS provider in subsection \ref{votingsection}. Below we explain each factor and the QoSs' associated with SDaaS services.
\begin{itemize}[leftmargin=*]
    \item \emph{Tangibles:} these refer to the appearance of a service. In the case of SDaaS, these refer to the QoS computed throughout the path composition.
    \begin{itemize}
        \item Delivery time: It is the total delivery time of the packages from a source to a destination in a skyway network. This includes the node times, i.e. the charging time and the waiting time due to sequential charging. The delivery time differs between different service providers based on the swarm's capabilities, techniques used, and composed path affected by partnerships.
        \item Energy consumed: It is the total power consumed during the delivery of the packages from a source to destination. Energy consumption may be an important preference to environmentalist and those interested in green, eco-friendly, and sustainable services.
        \item Cost: It is the total cost of the delivery from a source to a destination. It highly depends on the composed path, techniques used, and partnerships. Some consumers may prefer having a lower cost over a fast delivery.
    \end{itemize}
    \item \emph{Reliability:} this refers to the ability to perform the service accurately \cite{ladhari2009review}. In the case of SDaaS service, this refers to the successfulness of a delivery request, i.e. the ability of all drones to reach the destination within the expected time. This factor helps build trust between the consumer and the service provider.
    \item \emph{Responsiveness:} is the ability to provide prompt services. In SDaaS, the execution time of the composition algorithm differs based on the techniques used by the service provider. Hence, how quick the path will be computed is essential to deliver services quickly.
\end{itemize}

\subsection{SDaaS Providers Search Space Reduction}
We speculate to have many SDaaS providers in the future. Composing services for each provider would be a time extensive task. Therefore, we propose a search space reduction module to eliminate providers that will most probably not meet the preferences of the consumers. There are two main steps involved in the space reduction module described below:
\begin{itemize}[leftmargin=*]
    \item \textit{Filtering:} In this step, providers are filtered based on their capabilities to carry and deliver the consumer packages. For example, a request containing 2.5kg packages cannot be carried by a swarm whose members can carry a maximum of 1kg payload. In a similar manner, if the request includes five packages, then a swarm with four members cannot carry the request (assuming each drone can carry one package only). 
    \item \textit{Pruning:} The filtering step reduces the number of providers from $n$ to $m$, where $m \leqslant n$. The pruning step is intended to reduce the number of considered providers $u$ even further, where $u \leqslant m$. In this step, we consider the charging stations distribution across the network as shown in Fig. \ref{fig:heatmap_distribution}. We then plot a heatmap to represent the \textit{density} of each charging provider in a region $R_i$. The density is computed based on the number of stations within a region and the number of pads within each station. A heatmap for each charging station is generated as shown in the example in Fig. \ref{fig:heatmap_example}. As shown in the figure, we look into the regions that the swarm is most likely to visit on its way from the source to the destination. We compute the $t$ shortest paths across the heatmap regions from the source to the destination. As shown in Fig. \ref{fig:heatmap_example} the right most cells are less likely to be visited as they are far from the source and destination. For each $t$ path computed, we calculate an aggregated score of an estimate cost of visiting this path based on the partnership and the density. The intuition behind this step is that paths with higher partnership are most likely going to result in better composition. We take the best aggregated value from the $t$ paths to represent a provider value. The value of each provider takes into consideration the swarm capabilities as well, e.g. swarm size, battery capacity, etc. Equation \ref{scoreequation} shows how a provider's score is computed for a single path. 
    \begin{equation}
    \label{scoreequation}
         Path Score  = \left( \sum_{R=i}^{R=paths_R} R_{Density} \right) * S_{Capabilities}
    \end{equation}
    
    Now that each provider is associated with a score, the providers are sorted based on the score. Providers with the lowest $k$ scores are pruned out as shown in Fig. \ref{fig:framework}. This is due to the fact that these $SDaaS$ providers are unlikely to be good enough to compete for the request. The output of this module are the $u$ providers left that will compete in the next modules to serve the request.

    
    \begin{figure}[!ht]
    \begin{subfigure}{.5\textwidth}
      \centering
      \includegraphics[width=.75\linewidth]{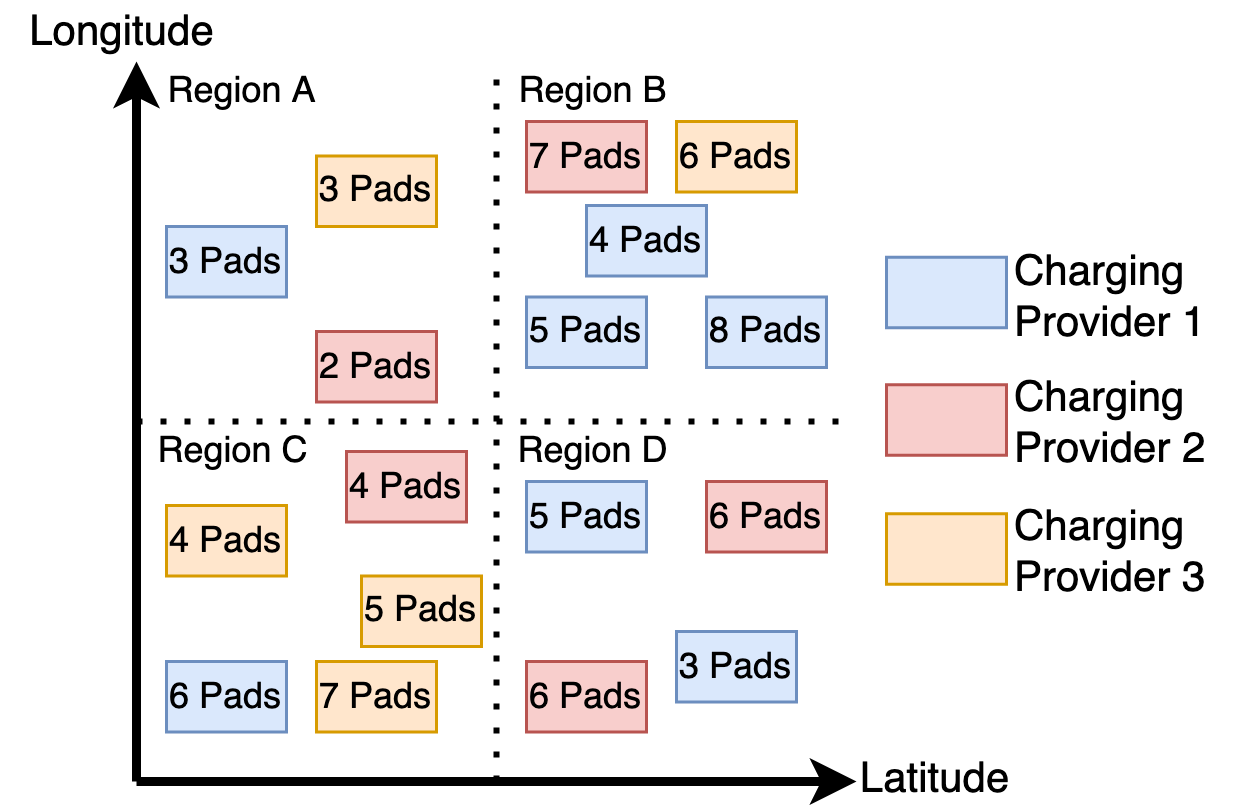}  
      \caption{Distribution of Charging Stations Owned by Different Providers}
      \label{fig:heatmap_distribution}
    \end{subfigure}
    \begin{subfigure}{.5\textwidth}
      \centering
      \includegraphics[width=.7\linewidth]{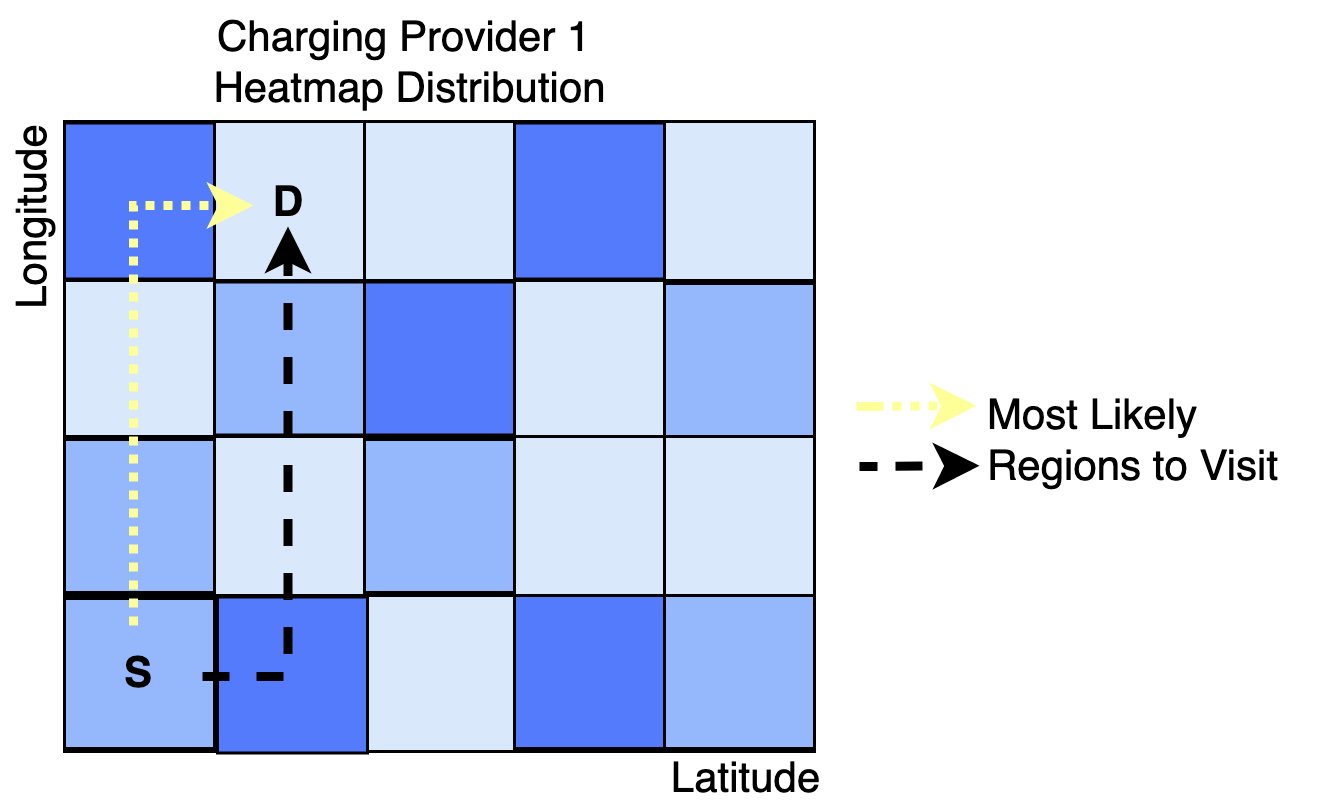}  
      \caption{Density Heatmap of Charging Provider 1 Across the Network}
      \label{fig:heatmap_example}
    \end{subfigure}
    \caption{Density Heatmap of Charging Providers}
    \label{fig:fig}
    \end{figure}

    
    
\end{itemize}

\subsection{SDaaS Composition}
\label{composition}
There are two main factors that control the composition of the optimal services from a source to a destination. First are the providers swarm capabilities and techniques. These include the size of the swarm, the payload capacities of the drones, the energy consumption rates of the drones, etc. Second are the consumers weighted preferences. For example, a consumer may most care about a cheap delivery and would put higher weight on the cost compared to the delivery time. As shown in Fig. \ref{fig:framework}, these two factors are the input for the weighted composition algorithm adapted from \cite{alkouz2020swarm}. 

There are several constraints that need to be addressed during the composition of the optimal path. Below, we describe the different constraints that exist only in swarm-based deliveries as opposed to single drones deliveries.
\begin{itemize}[leftmargin=*]
    \item \emph{In-swarm varied energy consumption:} The drones in a swarm would typically consume energy at different rates due to the different carried payloads, the selected formation, and the position of the drone in the swarm. The drones positions affect the consumption due to the drag and upwash/downwash forces of the neighbouring drones \cite{liu2022constraint}.
    
    \item \emph{Limited recharging pads:} As the number of drones in a swarm are typically large, there might be instances where the number of drones exceed the number of recharging pads at a node. In this case, the drones need to recharge sequentially increasing the waiting times and the overall delivery times.
    
    \item \emph{Constrained arrival time window:} As the swarm may split and disperse in the network in the case of a dynamic behaviour, the sub-swarms need to adhere to a limited arrival time window that meet the consumer's expectations \cite{alkouz2021service}.
    
    \item \emph{Preference-based vs capabilities-based composition:} The composition of the optimal services must cater for two needs. The first is the weighted preferences of the consumer. For example, if the consumer cares about the cost, the swarm may choose a path with partnerships to decrease the cost even if the path is longer or the number of available recharging pads is less than other paths. The second is the provider swarms capabilities. A provider for example may not have any partnership with any station and would need to compose the path regardless of the cost associated.
\end{itemize}
\begin{figure*}[t]
\centering
\includegraphics[width=0.9\linewidth]{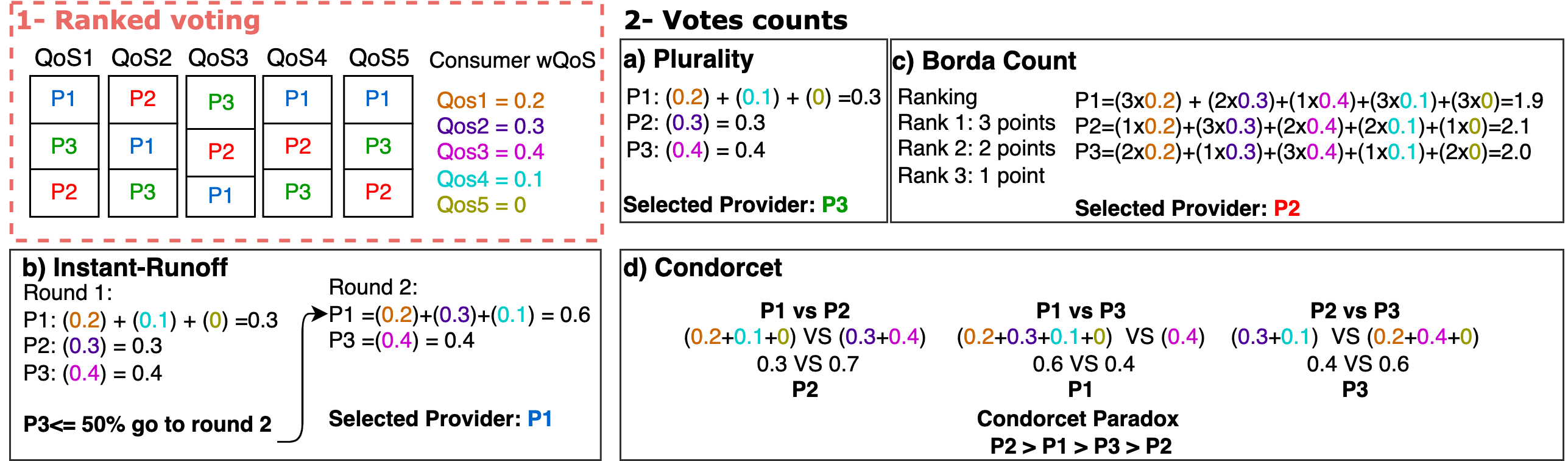}
\caption{Voting Count Systems}
\label{fig:votingMethods}
\end{figure*}

As described earlier, the composition of the path is delegated by the providers to the broker. The broker takes the capabilities of the provider and preferences of the consumer and tries to compose the optimal path given the techniques the provider can undertake described in subsection \ref{techniques}. In the composition, the swarm with drone $D$ at the source computes the potential to reach the destination directly using the shortest dijkstra's path. The potential here is the ability for all drones $d$ in $D$ to reach the destination successfully without consuming all it's energy. 
If the swarm is not able to reach the destination, it goes to the best neighboring node. The best nodes is the nodes that gives the best weighted QoS based on the consumer's preference using the following equation:

\begin{equation}
    node\_value=\sum_{i=1}^{n}\hat{wQoS}*\hat{pQoS}
\end{equation}

where wQoS is the normalized weight specified by the consumer for every QoS and pQoS is the normalized perceived value of the QoS if this node gets selected.

The neighbour with the best $node\_value$ gets selected and the swarm traverses to the node. In case the provider chooses a dynamic swarm behaviour technique, the swarm would split into two sub-swarms and the best two neighboring nodes are traversed to. The split may cause decreased waiting times due to sequential charging \cite{alkouz2020swarm}. If the provider selects a flexible formation technique, the swarm would change formation as the weather changes to consume less energy. Otherwise, if the provider uses a fixed formation and sequential composition, the execution times of the composition would reduce as there are less computations done \cite{alkouz2020swarm}\cite{alkouz2020formation}. If a cooperative behaviour technique is used, the swarm charges up to what takes it to the next node instead of charging the batteries fully. At every neighbour, the $pQoSs$ values are updated. For example, $pQoS_{dt}+=tt+ct+wt$, where $tt$ is the segment travel time, $ct$ is the nodes charging times, $wt$ is the waiting times in case of sequential charging. The swarm again tries to reach the destination directly, otherwise, it goes to the next best neighbouring node until it reaches the destination \cite{10.1145/3460418.3479289}. The output of this step is a set of perceived QoSs' for all providers. For example, the perceived QoSs' for the composed path for provider 1 are, $<p_1\_dt,p_1\_ec,p_1\_et,p_1\_c>$. This output will be used to recommend the best providers to carry-on this delivery in the next subsection.

\subsection{Voting-based SDaaS Provider Recommendation}
\label{votingsection}

The SDaaS composition results in a set of perceived QoS values for every provider. These values along with the consumer weighted QoS preferences are the input to the SDaaS provider recommendation module (see Fig. \ref{fig:framework}). We propose a voting-based provider recommendation mechanism to recommend the best SDaaS provider to a consumer's request.

The notion of voting is commonly used in political election settings. However, we propose to use voting to recommend the best provider for a request. The voters are the QoS metrics of SDaaS services, i.e. delivery time, cost, etc. Each QoS will rank the providers based on the perceived values ($pQoS$). For example, if provider 3 gave the best delivery time and provider 2 gave the second best delivery time then, $dt = {p3,p2,..,pu}$. The process is repeated for every QoS metric as shown in step 1 in Fig. \ref{fig:votingMethods}. Then, to aggregate the votes, several vote count systems are implemented. We propose to evaluate four vote count systems namely, Plurality, Instant Runoff, Borda Count, and Condorcet. We describe each method below:

\begin{itemize}[leftmargin=*]
    \item \textbf{Plurality:} also known as relative majority voting, is one of the most common and straight forward types of votings. In this method, each voter (QoS) is allowed to vote for one SDaaS provider; i.e. the provider with the highest perceived value is selected. The provider that polls more than any other candidate is the winner \cite{donovan2019self}.
    \item \textbf{Instant Runoff:} also known as ranked choice voting (RCV).
    Ballots are first counted for each voter's top choice. If one of the SDaaS providers get more than half of the votes based on the top choice, that provider is announced as the winner. If not, the SDaaS provider that received the least votes in the first round gets eliminated. If there are multiple minimums, a random of the minimums gets eliminated. The votes of those that selected the eliminated provider as their first choice have their votes added to the totals of their next choice. This process continues until one of the SDaaS providers get more than half of the total votes.
    \item \textbf{Borda Count:} similar to Instant Runoff, the voters (QoS) rank their choices based on $pQoS$. However, in this method, each rank has certain points associated with it \cite{lippman2013voting}. For example, rank 1 gets 3 points, rank 2 gets 2 points, etc. If SDaaS provider $i$ is ranked 1, two times and ranked 2, three times, then its total points is (3*2)+(2*3)=12. Once the points are counted for all SDaaS providers, the provider with the highest points get selected as the winner.
    \item \textbf{Condorcet:} is a head-to-head election between every pair of SDaaS providers \cite{gehrlein2001condorcet}. The counts by voters (QoS) for every pair is counted and the \emph{local} winner is the one with highest votes between the two. The process is repeated for every pair. The \emph{final global} winner is the SDaaS provider with highest wins in every local match. A Condorcet paradox would occur when the preferred providers are cyclic, i.e. the preferences can be in conflict with each other. Suppose the QoSs' prefer, for example, provider A over B, B over C, and yet C over A. In this case, a paradox happens.
\end{itemize}

For every method and before selecting the winner, the output of every round is multiplied by the weighted QoS ($wQoS$) given by the consumer's request. In this case, if the consumer prefers the delivery time QoS over the others and a certain provider performs best in other QoS, the provider will not be selected as a winner due to the weight. Fig. \ref{fig:votingMethods} illustrates an example using the different voting count systems. In step 1 the different QoS ranked the providers based on the perceived values after composition ($pQoS$) as described earlier. In step 2, the votes are aggregated using the different methods. Note how the count values are multiplied by the weights from the consumer's requests ($wQoS$). As shown from the example, each voting count method selects and recommends a different provider. In the Condorcet method, the paradox occurs where each provider is preferred over the other. In this case a random provider is recommended.

\section{Experiments and Results}

\begin{figure}[!ht]
    \begin{subfigure}{.5\textwidth}
      \centering
      \includegraphics[width=.8\linewidth]{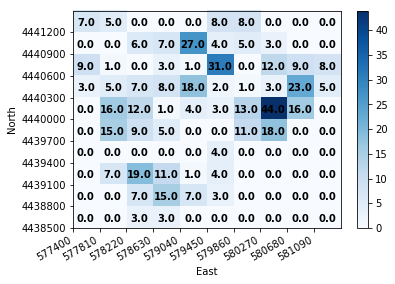}  
      \caption{Company A Heatmap Density Distribution of Charging Stations}
      \label{fig:company2}
    \end{subfigure}
    \begin{subfigure}{.5\textwidth}
      \centering
      \includegraphics[width=.8\linewidth]{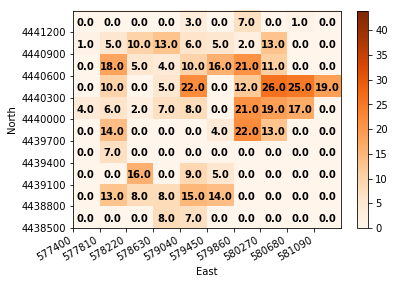}  
      \caption{Company B Heatmap Density Distribution of Charging Stations}
      \label{fig:company4}
    \end{subfigure}
    \caption{Dataset Density Distribution of Charging Stations}
    \label{fig:companies}
    \end{figure}

In this section, we evaluate the performance of the proposed framework. We conduct a set of experiments to evaluate the performance in terms of search space reduction cost, consumer satisfaction, and run-time efficiency. The dataset used in the experiments is an urban road network dataset from New York city. The data consist of a graph edge list of the city \cite{karduni2016protocol}. For the experiments we took a sub-network of 492 connected nodes to mimic a skyway network. Each node was allocated with different number of recharging pads and owned by different charging station providers randomly. We assigned the nodes to five different charging station providers. Figs. \ref{fig:company2} and \ref{fig:company4} represent a density heatmap distribution of charging stations of two different providers in the network using the UTM map system. The density value takes into consideration the number of stations and pads within each station. Since the dataset used is a real urban dataset, crowded and more dense areas are spotted in the heatmaps. The middle right regions are most likely CBD areas that are usually more crowded and would typically be more serviced by charging stations. The distribution mimics how a real world scenario of charging stations distribution would look like in a skyway network.

We then generate a set of 200 requests with a different source and destination nodes. For each request, we synthesize payloads with a maximum size of 10 packages and a maximum weight of 2.5 kg. For each request we generate a set of 40 providers with random capabilities and partnership tiers with different charging station providers. We run all the experiments using Python on a MacBook Pro, Apple M1 chip, 16 GB memory, and 8 cores.

We compute a satisfaction score to measure how much the recommended provider satisfies a consumer's request. The QoS satisfaction is computed using SERVQUAL satisfaction equation of weighted preferences \cite{parasuraman1988servqual} as shown below:
\begin{equation}
    \label{satisfactioneq}
    QoS_i \ Satisfaction = wQoS_i (pQoS_i - eQoS_i)
\end{equation}

where $wQoS_i$ is the weight that the consumer assigned for the importance of this QoS metric. $pQoS_i$ is the perceived actual value after composition. $eQoS_i$ is the expected quality of service. An example of $eQoS$ is Google Maps that gives the user an expected travel time from point A to B. In a similar manner, we compute $eQoS_i$ using the range of realistic expectations that is normally within the maximum and minimum values of $pQoS_i$ taking into consideration the consumers preferences. For example, if the possible range of $pQoS_{dt}$ for a certain request is between 15 and 50 minutes and the $wQoS_{dt}$ is 0.6, then the $eQoS_{dt}$ would be ((50-15)*0.6)+15=36. We normalize all the actual values before computing the satisfaction. The average of all QoS satisfaction scores represent the overall satisfaction. A satisfaction score of 0 means that the request is fully satisfied. A satisfaction score more than 0 means that the provider selected exceeds expectation. A satisfaction score less than 0 means that the provider selected is below expectation. 

\begin{figure}[!ht]
    \begin{subfigure}{.5\textwidth}
      \centering
      \includegraphics[width=.85\linewidth]{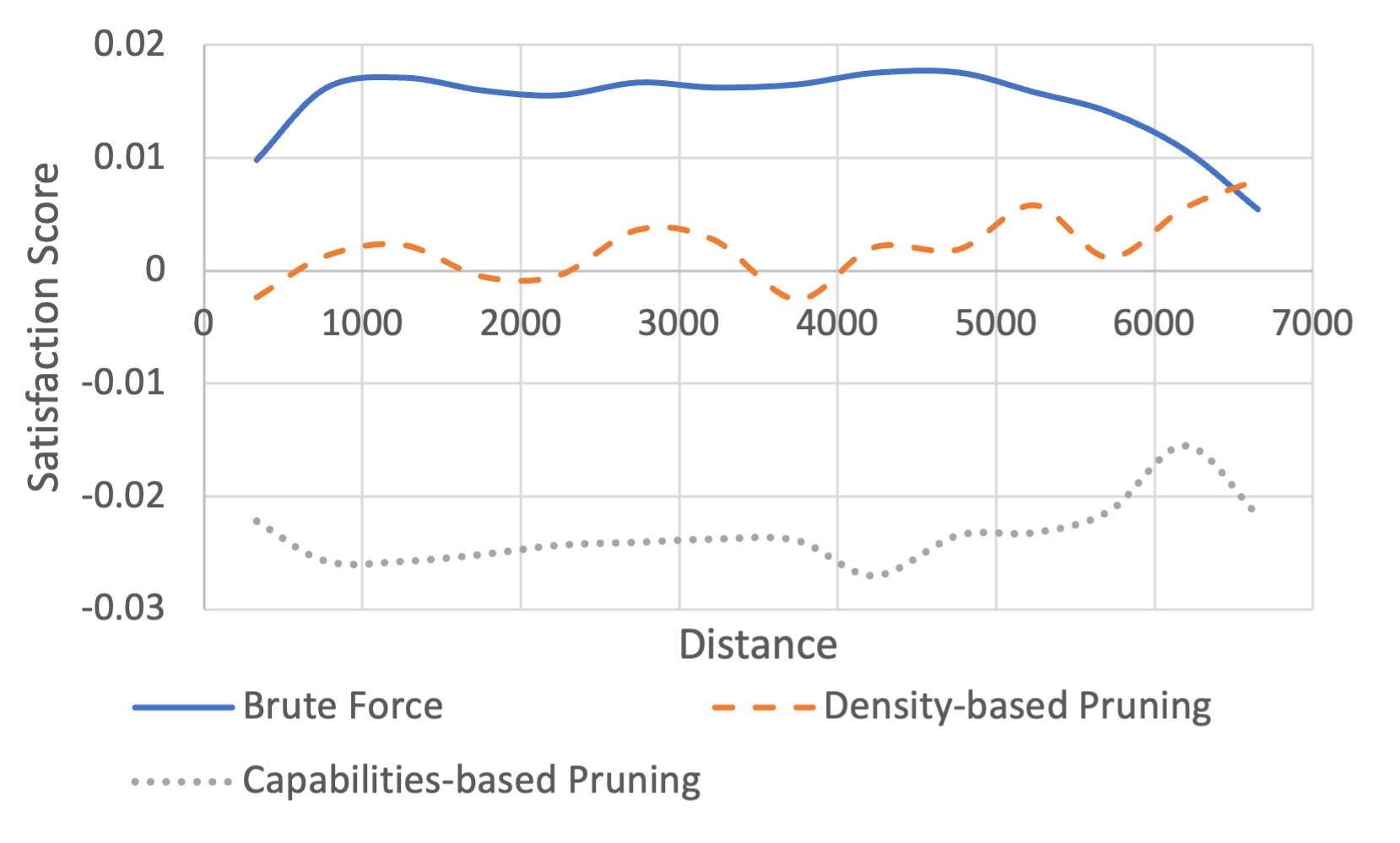}  
      \caption{Pruning Approaches Satisfaction Score}
      \label{fig:prunsat}
    \end{subfigure}
    \begin{subfigure}{.5\textwidth}
      \centering
      \includegraphics[width=.85\linewidth]{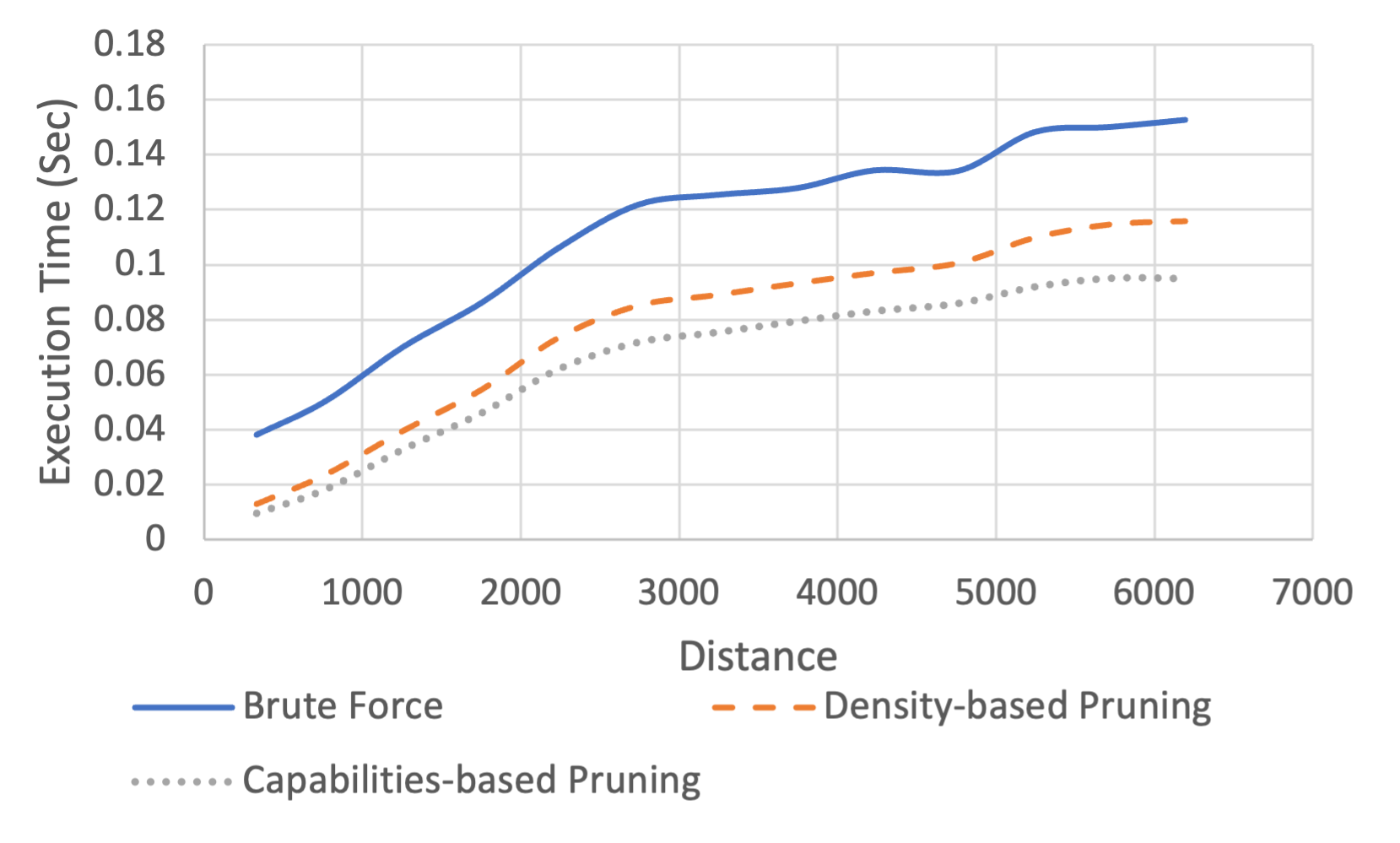}  
      \caption{Pruning Approaches Execution Time}
      \label{fig:prunet}
    \end{subfigure}
    \caption{Pruning Approaches Evaluation (Instant Runoff $k$=50\%)}
    \label{fig:pruning}
    \end{figure}

In the first experiment, we measure the performance of the density-based pruning approach against a brute force and a capabilities-based pruning approaches. In the brute force approach, no pruning happens. The path is composed and the voting-based recommendation takes place on all eligible providers after the filtering module. In the capabilities-based pruning approach, the $m$ providers left after the filtering module are given a score based on their capabilities, e.g. battery and speed. After sorting the providers according to their scores, the bottom $k$ percent is removed. Fig. \ref{fig:prunsat} represents the satisfaction scores of the different approaches. We fix the voting count method to Instant Runoff and the $k$ pruning value to 50\%. As shown in the figure, the brute force approach outperforms the other methods. This is because with pruning there is a risk of removing potentially better providers. However, brute force comes with the cost of execution time as shown in Fig. \ref{fig:prunet}. The pruning approaches cut the computation time significantly. Moreover, as the distance increases the difference becomes more significant as the composition algorithm becomes more costly for every provider. The density-based pruning approach resulted in an average of satisfied and above satisfied scores. This justifies that our proposed density-based pruning approach reduces the search space smartly and keeps providers that are mostly competitive. The capabilities-based pruning performed the worse as the pruning does not take into consideration the state of the network and the partnerships with charging providers.

\begin{figure}[!ht]
    \begin{subfigure}{.5\textwidth}
      \centering
      \includegraphics[width=.85\linewidth]{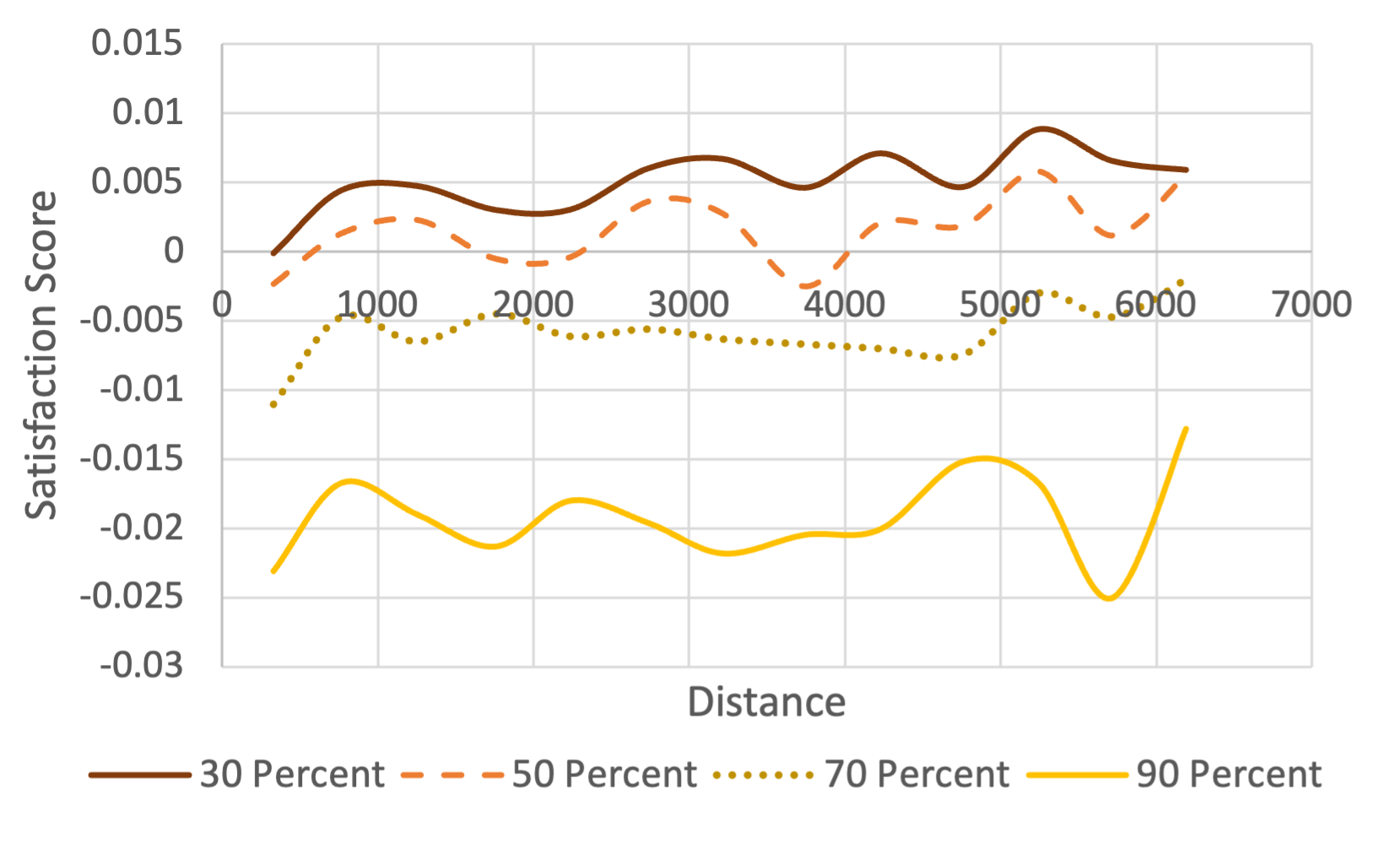}  
      \caption{Satisfaction Score with Varying $k$ Percent }
      \label{fig:ksat}
    \end{subfigure}
    \begin{subfigure}{.5\textwidth}
      \centering
      \includegraphics[width=.85\linewidth]{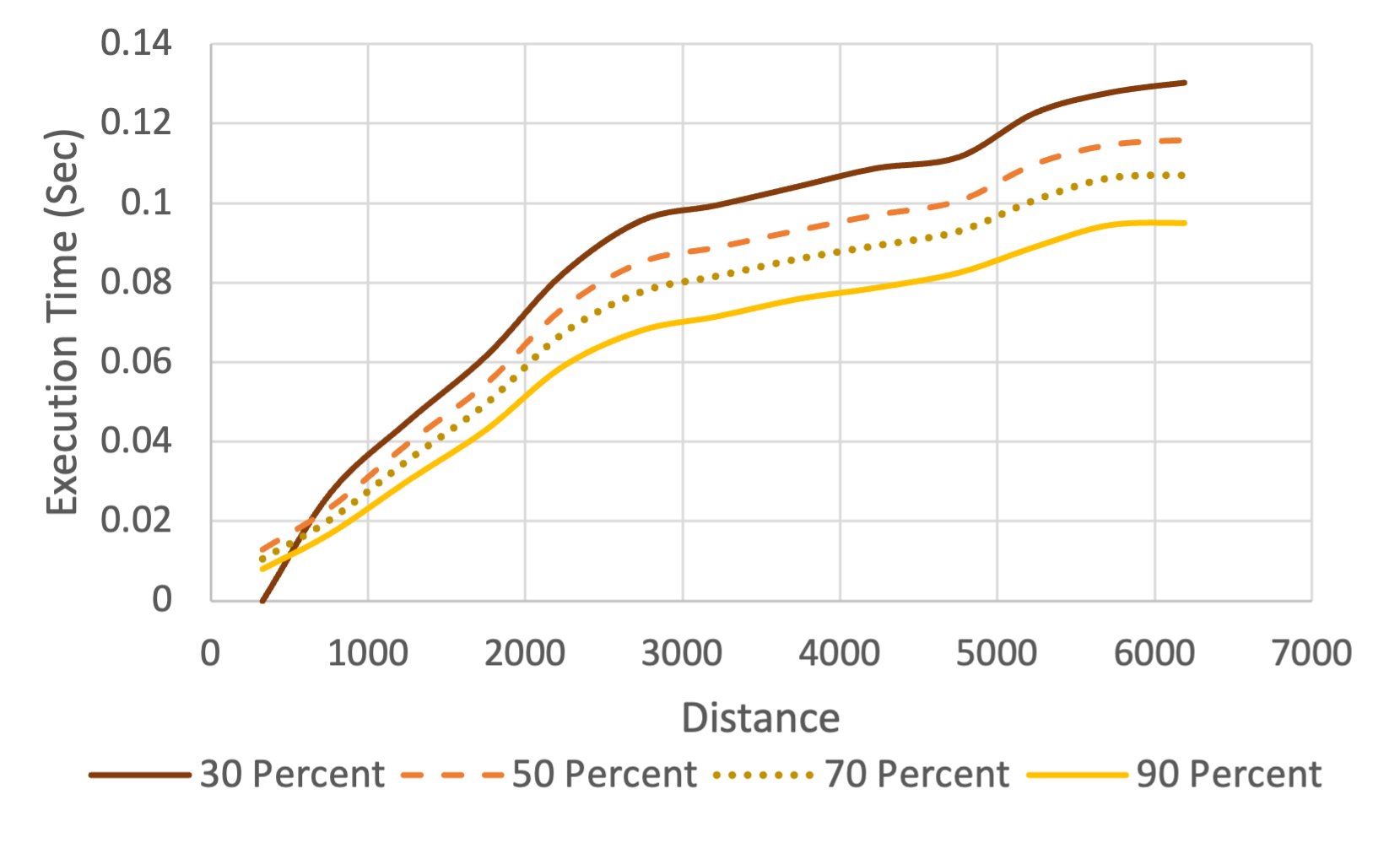}  
      \caption{Execution Time with Varying $k$ Percent}
      \label{fig:ket}
    \end{subfigure}
    \caption{Varying $k$ Percent in Density-based Pruning (Instant Runoff)}
    \label{fig:k}
    \end{figure}

In the second experiment, we evaluate the loss in the accuracy of the recommendation as the value of $k$ percent varies. Since the number of providers after filtering vary for each request, we prune the bottom $k$ percent instead of bottom $k$ providers. The higher the $k$ percent value the more providers are pruned out. We fix the voting count method to Instant Runoff. Fig. \ref{fig:ksat} shows that more loss occurs as more providers are pruned. This is an expected behaviour as the chances of recommending the best provider reduces since less number of them proceed to the voting-based recommendation step. On the other hand, the higher the $k$ percent value the lower the computation time as shown in Fig. \ref{fig:ket}. This is because the composition and voting-based recommendation consider less number of providers. Depending on the application, the trade off between accuracy in satisfying the consumers and run-time efficiency could be extracted from these graphs. For example, in a case of delivering high priority items in emergencies, the recommendation of an optimal provider that would satisfy the request need is more important than the computation cost.  Even with the lowest and most accurate $k$ percent value, i.e. 30\%. The execution is still significantly lower than the brute force in Fig. \ref{fig:prunet}.
    
\begin{figure}[!ht]
    \begin{subfigure}{.5\textwidth}
      \centering
      \includegraphics[width=.85\linewidth]{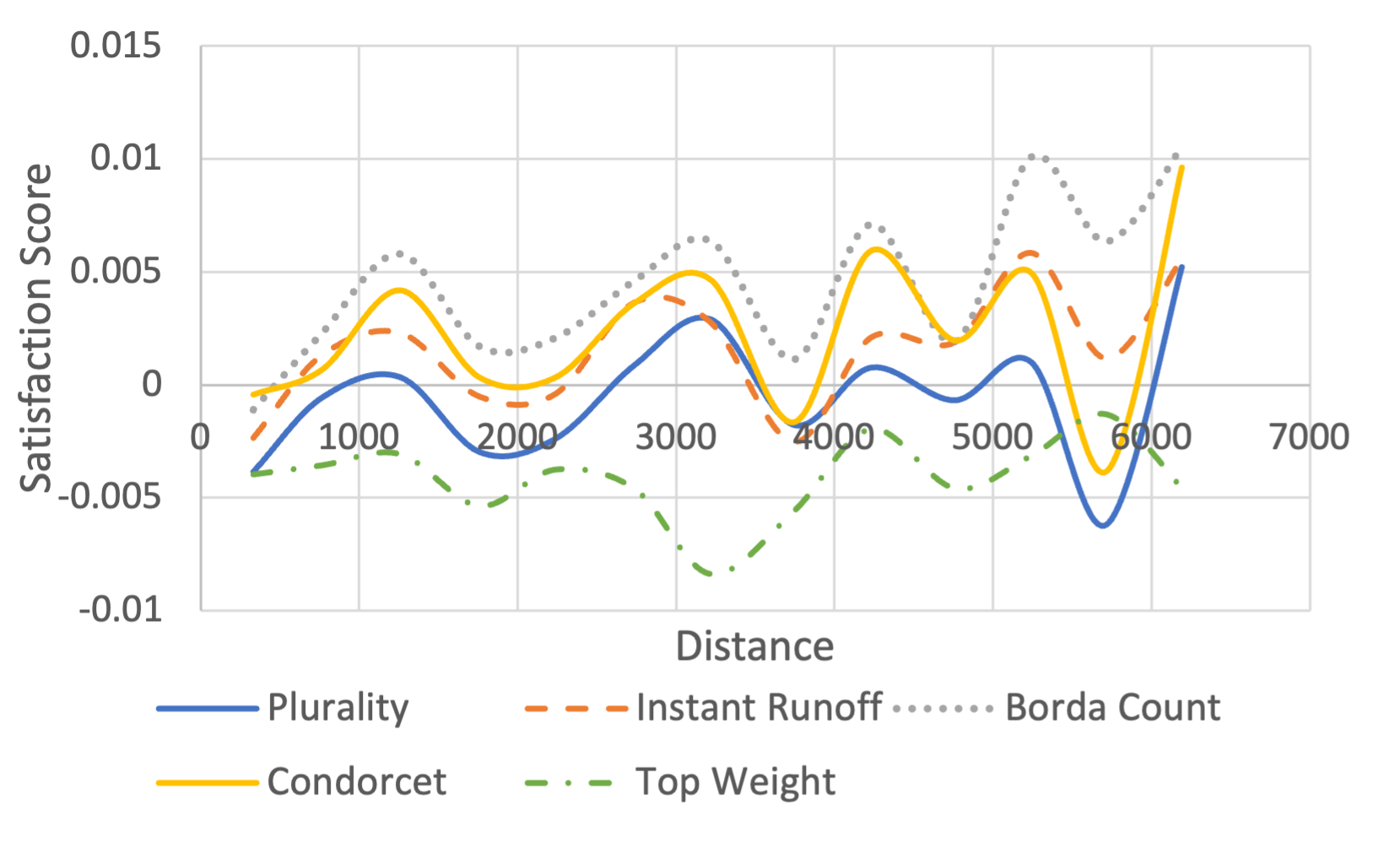}  
      \caption{Satisfaction Score with Different Vote Count Systems }
      \label{fig:votingsat}
    \end{subfigure}
    \begin{subfigure}{.5\textwidth}
      \centering
      \includegraphics[width=.85\linewidth]{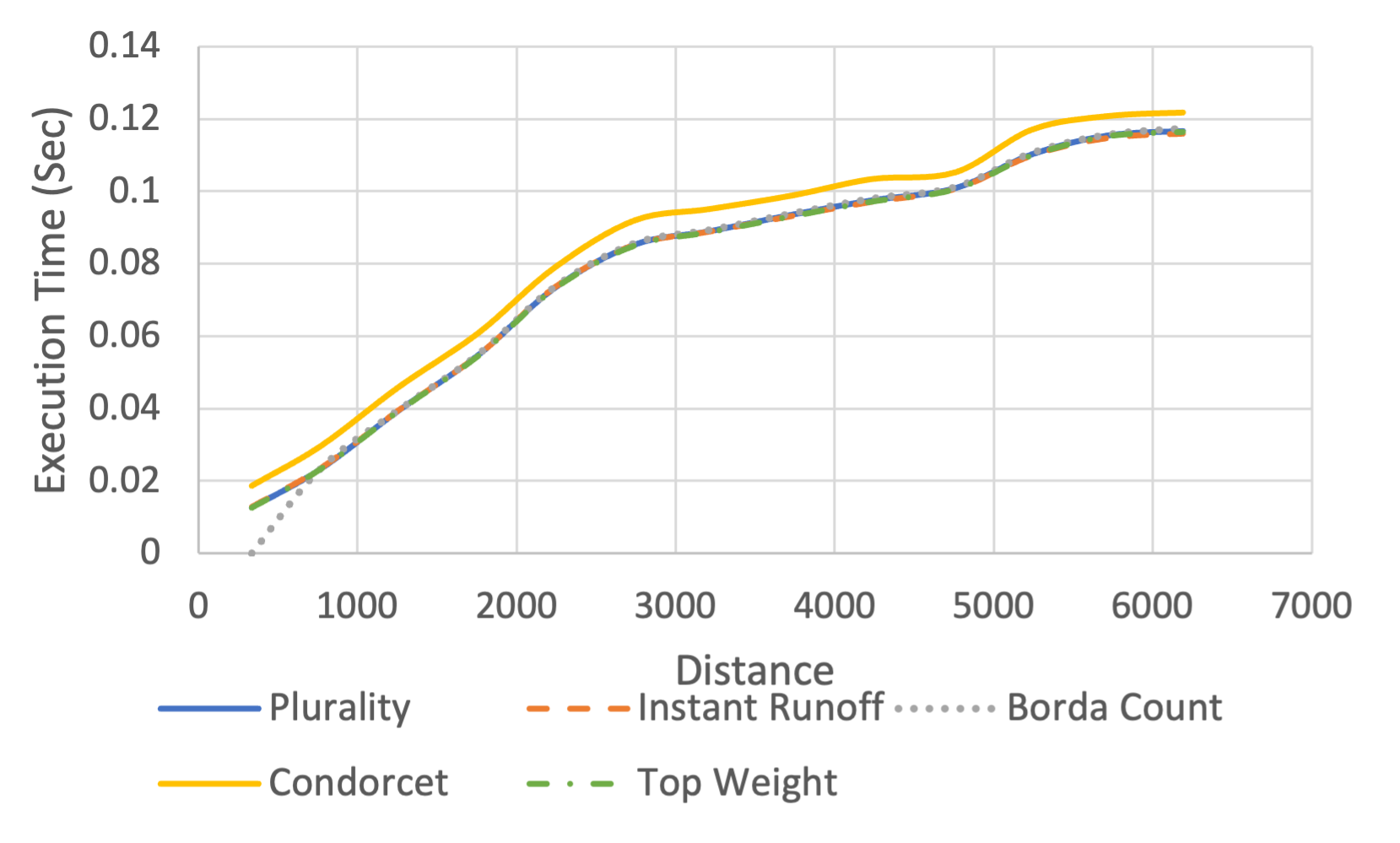}  
      \caption{Execution Time with Different Vote Count Systems}
      \label{fig:votinget}
    \end{subfigure}
    \caption{Vote Count Systems Evaluation ($k$=50\%)}
    \label{fig:votingresults}
    \end{figure}

In the last experiment, we measure the effectiveness of the vote count systems. These systems are evaluated against a Top Weight method where the provider with the best perceived value ($pQoS_i$) of the highest weighted QoS ($wQoS_i$) is recommended. For example, if the consumers puts the highest weight on the delivery time QoS, then the provider that returns the shortest delivery time is recommended. Fig. \ref{fig:votingsat} present that satisfaction scores with the different methods. We fix the $k$ pruning value to 50\%. As shown in the graph, all proposed vote count systems perform better than the Top Weight method. This is because the SDaaS provider recommendation problem is a multi criteria decision making (MCDM) problem. The recommendation problem lies under MCDM because the provider needs to be selected based on multiple criteria according the $wQoS$. Therefore, a simple recommendation of a provider based on a single criteria would not be sufficient as shown in the Top Weight approach in Fig. \ref{fig:votingsat}. In the case of SDaaS providers recommendation, the Borda Count method outperforms the others. This is because in Borda Count, the decision is not only made based on the weights of every QoS but also the rank of each provider (see Fig. \ref{fig:votingMethods}). For example, if three out of five QoS choose provider 1 as their first ranked and the other two QoS rank provider 1 in the fith place. Then, provider 2 with two first ranks and three second rank would be recommended. The Borda Count system has the ability to choose different candidates, rather than the candidate that is preferred by the majority. The execution time of all voting methods is similar except the Condorcet system that takes a slightly longer time as shown in Fig. \ref{fig:votinget}. This is because in Condorcet, a head to head match takes place between all the combination of providers which needs more computation time.
    
\section{Conclusion}
We proposed a preference-based recommendation framework of swarm-based drone delivery services based on the consumers preferences. We presented the concept of charging station providers partnering with swarm delivery service providers. A density-based pruning technique was presented to reduce the search space of potential swarm services providers in order to satisfy the preferences of the consumers. A weighted services composition method was presented, which takes into account the providers' capabilities, as well as the consumers' preferences, in order to determine the best next service. We developed a voting-based recommendation of suppliers who best meet the needs and preferences of the consumers. We conducted a series of experiments to assess the framework's efficiency in terms of consumer satisfaction, run-time, and search space reduction cost. Experimental results show that proposed density-based pruning approach cuts the computation time significantly while satisfying consumers needs. Moreover, for SDaaS services recommendation, the Borda Count voting outperforms the other methods in terms of consumers satisfaction.

\section*{Acknowledgment}

This research was partly made possible by LE220100078 grant from the Australian Research Council. The statements made herein are solely the responsibility of the authors.




%
\bibliographystyle{ieeetr}
\bibliography{icws}

\end{document}